\documentclass{article}

\usepackage{arxiv}
\usepackage{subfigure}

\usepackage[utf8]{inputenc} 
\usepackage[T1]{fontenc}    
\usepackage{url}            
\usepackage{booktabs}       
\usepackage{amsfonts}       
\usepackage{nicefrac}       
\usepackage{lipsum}		
\usepackage{graphicx}
\usepackage{natbib}
\usepackage{doi}
\usepackage{makecell}
\usepackage[table]{xcolor}
\usepackage{longtable}
\usepackage{fancyhdr}

\usepackage{multirow}
\usepackage{caption}
\usepackage{enumitem}
\usepackage{multicol} 
\usepackage{lipsum} 

\usepackage{array}

\setlist[itemize]{leftmargin=10pt} 
\usepackage{listings}
\usepackage{xcolor}
\definecolor{codegreen}{rgb}{0,0.6,0}
\definecolor{codegray}{rgb}{0.5,0.5,0.5}
\definecolor{codepurple}{rgb}{0.58,0,0.82}
\definecolor{backcolour}{rgb}{0.95,0.95,0.92}

\lstdefinestyle{mystyle}{
    backgroundcolor=\color{backcolour},   
    commentstyle=\color{codegreen},
    keywordstyle=\color{magenta},
    numberstyle=\tiny\color{codegray},
    stringstyle=\color{codepurple},
    basicstyle=\ttfamily\footnotesize,
    breakatwhitespace=false,         
    breaklines=true,                 
    captionpos=b,                    
    keepspaces=true,                 
    numbers=left,                    
    numbersep=5pt,                  
    showspaces=false,                
    showstringspaces=false,
    showtabs=false,                  
    tabsize=2
}

\title{JT-SAFE-V2: SAFETY-BY-DESIGN FOUNDATION MODEL WITH WORLD-CONTEXT DATA}

\author{\textbf{Junlan Feng}\thanks{Corresponding author:~\texttt{fengjunlanit@chinamobile.com}}
        \mdseries~~~~Fanyu Meng
        ~~~~Chong Long
        ~~~~Pengyu Cong 
        ~~~~Duqing Wang
        ~~~~Yan Zheng \\[0.2em]
        ~~~~Yuyao Zhang
        ~~~~Xuanchang Gao
        ~~~~Ye Yuan
        ~~~~Yunfei Ma  
        ~~~~Zhijie Ren
        ~~~~Fan Yang \\[0.2em]
        ~~~~Na Wu
        ~~~~Di Jin
        ~~~~Chao Deng
        \\[0.3em]
        \textit{JIUTIAN Research} \\[0.3em]
}

\date{}

\renewcommand{\headeright}{Technical Report}
\renewcommand{\shorttitle}{JT-Safe-V2}
\begin{document}
\makeatletter
\setlength{\headheight}{25.75795pt}
\makeatother

\maketitle
\vspace{-18pt}

\begin{abstract}
We introduce JT-Safe-V2, a large language model designed to advance the safety and trustworthiness of foundation models, extending our previous JT-Safe model toward a more comprehensive safety-by-design paradigm. JT-Safe-V2 emphasizes the joint optimization of general intelligence and safety-by-design through several key innovations: enriching pre-training data with contextual world knowledge, high-certainty pre-training procedures, and safety strengthening post-training mechanisms for enterprise-oriented agentic capabilities. Building on these safety-enhanced foundation models, we propose Safe-MoMA (Safe Mixture of Models and Agents), a framework that enables traceable and efficient inference through the orchestrated deployment of multiple models and agents. Extensive evaluations demonstrate that JT-Safe-V2 achieves state-of-the-art performance across both general intelligence and safety benchmarks. Moreover, Safe-MoMA reduces inference costs by more than 30\% compared to using the largest standalone model baseline while maintaining comparable performance. To facilitate future research on safety-by-design foundation models, we publicly release the post-trained JT-Safe-V2-35B model checkpoint.

\end{abstract}


\section{Introduction}

Large language models (LLMs) have rapidly evolved into a foundational technology for artificial intelligence, demonstrating remarkable capabilities in language understanding, reasoning, code generation, and decision support \citep{deepseekv3.2_2025,guo2025deepseek}. Advances in large-scale pre-training and instruction alignment have enabled these models to perform a wide variety of complex tasks with minimal task-specific supervision \citep{openai2023gpt4,team2024gemini}. As a result, LLMs are increasingly deployed in real-world applications ranging from enterprise knowledge management and software engineering to intelligent agents and autonomous decision systems \citep{gpt5system}.

However, as LLMs become more deeply integrated into critical infrastructures and industrial workflows, safety and trustworthiness have emerged as fundamental challenges~\citep{2407.04295,huang2025trustworthiness}. Models may generate factually incorrect content, exhibit overconfident hallucinations, or produce unsafe responses under adversarial prompts~\citep{2408.07663,2504.15585,bommasani2021foundation}. These behaviors stem partly from the statistical nature of language modeling and the limitations of large-scale training data, which may contain outdated information, conflicting knowledge, or implicit biases~\citep{2102.02503}. When such models are deployed in high-impact environments, these issues can lead not only to degraded user trust but also to operational risks in domains such as finance, healthcare, public governance, and enterprise decision support systems. Ensuring reliable and safe model behavior has therefore become a central concern in the development of modern foundation models~\citep{2502.06872,2503.09648}.

A growing body of research has attempted to mitigate these risks through post-training alignment techniques, including reinforcement learning from human feedback (RLHF)~\citep{ouyang2022training}, safety classifiers~\citep{2412.17686,2501.16534}, and external guardrail systems~\citep{2502.05206,2505.03574}. These methods aim to guide model behavior toward safer responses by incorporating human preferences and safety constraints during fine-tuning or inference. While they can partially reduce harmful outputs, such approaches typically function as external control layers applied after the core model has already been trained.

Recent analyses suggest that many reliability problems originate earlier in the model development pipeline. Two factors are particularly influential: the structure and quality of pre-training data, and the learning paradigm based on next-token prediction~\citep{2202.03629}. Web-scale datasets inevitably contain noise, misinformation, and incomplete representations of real-world knowledge. When models learn statistical correlations from such data without sufficient contextual grounding, they may internalize uncertainty and reproduce it during generation~\citep{huang2025trustworthiness}. Moreover, predictive objectives that focus solely on token likelihood can encourage models to produce plausible yet unsupported statements when the underlying knowledge is ambiguous. These characteristics highlight the limitations of treating safety purely as a post-training problem.

These observations motivate a shift from reactive safety mechanisms toward safety-by-design foundation models, where safety and trustworthiness are treated as intrinsic design objectives throughout the entire model lifecycle.

In this work, we introduce JT-Safe-V2, a new generation of safety-oriented large language models designed to jointly optimize general intelligence and safety properties. Instead of relying solely on post-training corrections, JT-Safe-V2 integrates safety considerations directly into data construction, training procedures, and alignment mechanisms, aiming to improve reliability while maintaining strong reasoning capabilities.
Building on these models, we further propose Safe-MoMA (Safe Mixture of Models and Agents), a scalable inference framework for enterprise-oriented agentic systems. Safe-MoMA dynamically orchestrates heterogeneous models and agents according to task complexity and resource constraints, enabling traceable and cost-efficient execution.
Extensive evaluations show that JT-Safe-V2 achieves strong performance across both capability and safety benchmarks, while Safe-MoMA significantly reduces inference costs without sacrificing task performance.

Finally, to support future research on safety-by-design foundation models, we release the post-trained JT-Safe-V2-35B model checkpoint together with detailed technical documentation.

In summary, this work makes the following contributions:

\begin{itemize}
    \item \textbf{Safety-by-design foundation models.} We introduce JT-Safe-V2, a series of large language models designed to jointly optimize capability and safety. Unlike conventional approaches that treat safety as a post-training constraint, JT-Safe-V2 incorporates safety considerations throughout the entire model lifecycle, including data construction, training procedures, and post-training alignment. This design improves model reliability while preserving strong reasoning and problem-solving capabilities.
    \item \textbf{Context-aware pre-training data.} We develop the Data with World Context (DWC) corpus, which augments large-scale text corpora with structured contextual signals such as temporal metadata, domain classifications, source credibility indicators, and safety attributes. This contextual grounding improves factual reliability and helps mitigate uncertainty inherited from raw web-scale data.
    \item \textbf{Safe-MoMA inference framework.}We propose Safe-MoMA (Safe Mixture of Models and Agents), a scalable inference framework that dynamically orchestrates heterogeneous models and agents according to task complexity and resource constraints. Safe-MoMA enables traceable execution and reduces inference cost by 42\% compared with a largest-model baseline while maintaining competitive performance.
\end{itemize}

\section{Design Principles of Safety-by-Design Foundation Models}

The design of JT-Safe-V2 is motivated by the observation that many safety and reliability issues in large language models originate from earlier stages of model development rather than solely from inference-time behavior. Traditional safety mitigation approaches often rely on external moderation mechanisms, but such strategies treat safety as an auxiliary component rather than an intrinsic property of the model.

To address this limitation, JT-Safe-V2 adopts a safety-by-design paradigm, where safety and trustworthiness are incorporated throughout the entire lifecycle of model development. This paradigm includes four core design principles.

First, data-centric trustworthiness. High-quality training data forms the foundation of reliable models. Instead of relying solely on large-scale web data, JT-Safe-V2 introduces contextualized datasets enriched with structured metadata, enabling models to interpret textual information within real-world contexts.

Second, high-certainty training procedures. The training pipeline integrates data governance mechanisms, structured knowledge augmentation, and uncertainty reduction strategies to ensure that the model learns stable knowledge representations rather than memorizing noisy patterns.

Third, safety-aware alignment mechanisms. Post-training alignment is designed not merely to constrain model outputs but to embed safety constraints directly into the model’s decision-making processes.

Finally, safe and efficient inference architectures. Modern AI systems increasingly rely on multiple models and agents collaborating to solve complex tasks. JT-Safe-V2 therefore integrates Safe-MoMA, a multi-model orchestration framework that balances capability, safety, and computational efficiency.
Together, these principles form the conceptual foundation for building trustworthy foundation models.

\section{Data with World Context (DWC)}
\subsection{Motivation for Contextualized Training Data}
Modern large language models are typically trained on massive corpora consisting of web pages, books, and code repositories. While these datasets contain enormous amounts of textual information, they fundamentally represent flattened textual data, where important contextual signals are often omitted or implicitly assumed.

In many cases, correct interpretation of textual information requires knowledge about temporal context, domain background, source reliability, and author intent. For example, a sentence such as “The president announced a new policy yesterday” lacks critical information such as the country, time period, and political context. Human readers naturally resolve these ambiguities using background knowledge, but language models trained purely on token-level statistics often struggle to infer the correct context.

As a result, traditional training datasets introduce several structural challenges:
\begin{itemize}
	\item \textbf{Factual ambiguity.} Raw textual data often lacks explicit temporal and spatial anchors, making it difficult for models to correctly associate events and entities.
	\item \textbf{Fragile reasoning.} Models trained on large-scale text corpora may learn statistical correlations rather than causal structures, leading to unstable reasoning behavior.
	\item \textbf{Passive learning.}The next-token prediction paradigm encourages pattern replication but does not explicitly capture the intent, learning objectives, or cognitive structure embedded in textual information.
\end{itemize}
%
%

These limitations suggest that improving the quality and structure of training data is a critical step toward building safer and more trustworthy foundation models.

To address these issues, we introduce Data with World Context (DWC), a contextualized data construction framework that enriches textual data with structured knowledge signals reflecting real-world contexts. Unlike traditional datasets that treat text as isolated sequences of tokens, DWC integrates contextual annotations that capture factual anchors, reasoning structures, and cognitive intent.

Importantly, DWC is designed as a shared data infrastructure across the entire training pipeline, supporting not only pre-training but also supervised fine-tuning and reinforcement learning stages. By providing consistent contextual signals throughout training, DWC improves both the reliability of model knowledge and the stability of alignment processes.

\subsection{Evolution of DWC from JT-Safe to JT-Safe-V2}
The concept of \textbf{Data with World Context (DWC)} was first introduced in the JT-Safe~\citep{2510.17918} framework to enrich pre-training data with contextual attributes such as document metadata, domain classifications, and source information. By augmenting textual corpora with structured contextual signals, DWC improves the grounding of textual knowledge and helps models interpret information within a clearer real-world context. However, the original version of DWC primarily focused on external metadata augmentation, providing contextual information about the environment in which a document was created rather than the internal structure of the knowledge contained in the text itself.

In JT-Safe-V2, the DWC framework undergoes a substantial upgrade that expands contextual enhancement from external attributes to internal cognitive structures. This upgrade can be summarized along three key dimensions.

First, the enhancement scope extends from external attributes to the internal textual structure of knowledge. Earlier DWC annotations mainly captured metadata such as publication time, source information, and domain taxonomy. The upgraded framework instead emphasizes the structural organization of information within the text, enabling models to better capture how knowledge is constructed and related across sentences and passages.

Second, the annotation architecture evolves from a single factual layer into a unified three-layer framework consisting of factual, logical, and cognitive annotations. The factual layer anchors objective knowledge elements, the logical layer reveals reasoning relationships embedded in the text, and the cognitive layer captures the intent and learning value underlying the information.

Third, the representation of contextual information shifts from discrete metadata labels to structured knowledge units. Instead of storing contextual attributes as independent key–value pairs, the new DWC framework introduces structured representations—such as arrays of knowledge points—that decompose long documents into modular knowledge components. This representation allows models to access and organize knowledge at a finer granularity.

Through these upgrades, DWC transitions from a contextual metadata enhancement method into a cognitive data infrastructure capable of supporting advanced reasoning, knowledge organization, and learning-oriented information processing.

\begin{figure}[ht]
	\centering
	\includegraphics[width=15cm]{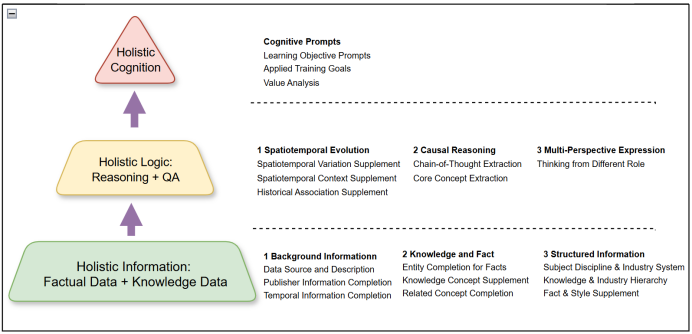}
	\caption{The three-layer annotation architecture of DWC.}
	\label{fig:data_processing_pipeline}
\end{figure}

\subsection{Annotation Architecture and Data Structure}
\subsubsection{Three-layer annotation architecture}
The upgraded DWC framework introduces a structured three-layer annotation architecture designed to enrich textual data with multiple levels of contextual information. Rather than treating documents as isolated text sequences, the architecture organizes contextual signals into complementary layers that describe factual grounding, reasoning structures, and cognitive intent. This layered design allows models to access different dimensions of knowledge during training and improves the interpretability of the data representation.

\textbf{Factual Layer.}The factual layer anchors textual information with explicit knowledge signals derived from the real-world context in which the document exists. Typical annotations include named entities, temporal attributes, geographic references, domain classifications, and document source information. These attributes provide concrete grounding signals that help models associate textual statements with identifiable knowledge elements. By explicitly encoding such contextual attributes, the factual layer reduces ambiguity in textual information and improves the stability of knowledge representations learned during pre-training.

\textbf{Logical Layer.}The logical layer captures reasoning structures embedded within the text. Instead of representing knowledge solely as independent facts, this layer exposes relationships between information units. Typical annotations include causal relations, temporal sequences, conditional dependencies, and multi-step reasoning chains.For datasets involving complex reasoning tasks, additional annotations such as explicit reasoning traces or chain-of-thought structures may be incorporated. These annotations make implicit reasoning patterns observable during training and help models learn how knowledge components are logically connected within a document.

\textbf{Cognitive Layer.}The cognitive layer describes the informational intent and learning value associated with a document. While the factual and logical layers focus on knowledge content and structure, the cognitive layer provides signals about how the information is intended to be interpreted or used. Typical annotations include reading purpose, learning objectives, value perspectives, and intended audience level.These signals guide models to interpret information not only at the surface semantic level but also in terms of knowledge utility and educational value. In practical settings, such annotations can help models adapt responses to different knowledge contexts, such as educational explanation, factual summarization, or analytical reasoning.

\subsubsection{Data Structure Representation}

To support efficient processing during training, each DWC data sample is represented using a structured JSON format that separates raw textual content from contextual annotations. This design allows the model training pipeline to independently access textual data and contextual signals.


The text field stores the original document content, while the meta field contains structured contextual annotations derived from the three-layer architecture. These may include domain classifications, temporal attributes, safety signals, reasoning hints, and structured knowledge representations.

\subsubsection{Knowledge Unit Representation}

For knowledge-intensive datasets, the upgraded DWC framework further introduces knowledge unit representations that decompose long documents into modular information components. Instead of treating an entire document as a single training instance, the framework organizes information into arrays of knowledge points that represent atomic concepts or statements within the text.

This representation improves the informational density of training samples and enables models to access knowledge at a finer granularity. It also facilitates structured reasoning over document content, allowing models to identify relationships among knowledge units rather than relying solely on sequential token patterns.

Through this multi-layer annotation architecture and structured data representation, DWC evolves from a contextual metadata enhancement mechanism into a knowledge-aware data infrastructure capable of supporting advanced reasoning, structured knowledge organization, and learning-oriented model training.

\section{Safety-Centric Pre-training}
\subsection{High-Certainty Pre-training Procedures}
In traditional large language model pre-training, optimization often faces significant uncertainty due to heuristic learning rate schedules, such as cosine decay. These schedules are highly dependent on the estimated number of training steps and can cause instability when training is extended or when high-quality data with different distributions is introduced.

To overcome these limitations and ensure the model’s representation capacity increases monotonically, we adopt a high-certainty pre-training procedure. The core of this approach is to decouple the parameter optimization process from learning rate annealing, replacing heuristic strategies with deterministic post-hoc weight fusion. The procedure works as follows:

Deterministic Parameter Space Exploration: After an initial linear warm-up, the learning rate is fixed at a high constant value, allowing the model to perform broad parameter space exploration. During this stable stage, a larger optimization step size is maintained, ensuring the optimization trajectory remains global and not prematurely converging into local optima. The model checkpoints are saved at fixed intervals, preserving the optimization path.

Offline Decay Fitting and Optimal Weight Selection: Instead of using explicit learning rate decay during training, this approach simulates the decay process offline by averaging historical checkpoints using diversified weight distribution strategies. These models are then evaluated using core validation sets, and the model with the best generalization performance is selected as the final representation for the current training stage.

This procedure significantly enhances training stability, reducing optimization uncertainty while ensuring robust performance across varying training stages.

\begin{figure}[ht]
	\centering
	\includegraphics[width=17cm]{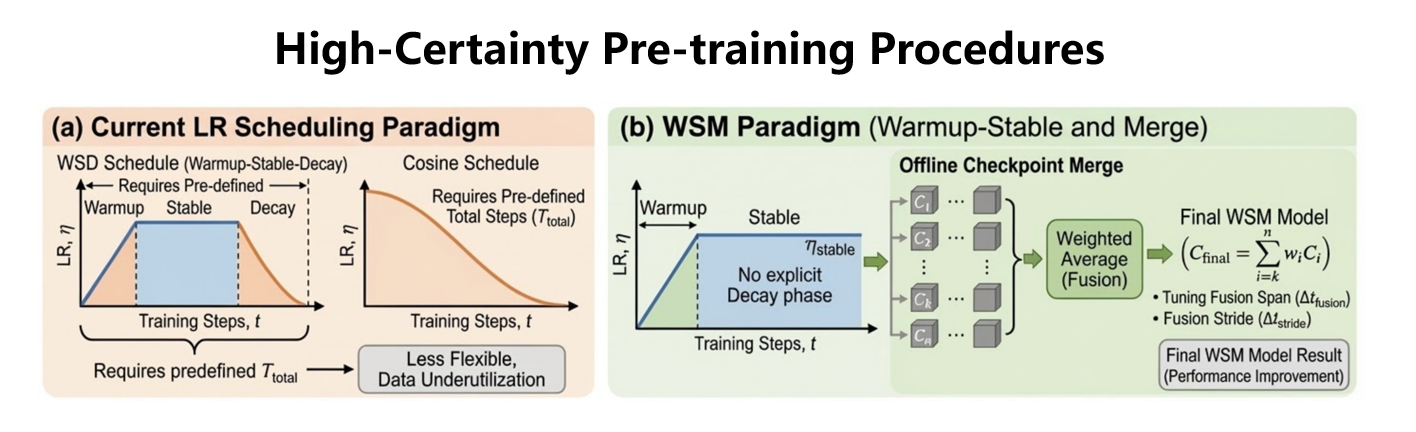}
	\caption{High-Certainty Pre-training Procedures.}
	\label{fig:pretrain_fig}
\end{figure}

\subsection{Context-aware Data Integration for DWC}
Incorporating DWC data into the pre-training pipeline requires modifying both the data format and loss calculation to ensure the model effectively leverages the contextual metadata.The key to integrating DWC data is to prepend the contextual metadata to the document text and use special tokens to mark the boundaries of this metadata. This allows the model to receive explicit signals about the document’s source, domain, release time, and credibility, which guide the model in understanding the contextual meaning of the text.

Additionally, tokens corresponding to the metadata are excluded from the loss computation, ensuring that the model’s language generation is focused solely on the text content while still conditioning on the additional contextual information.This mechanism ensures that DWC data enriches the model’s understanding without affecting its core language generation objective, improving both model stability and the interpretability of generated content.

One of the most significant challenges in large-scale continual pre-training is the issue of catastrophic forgetting—where the model forgets previously learned information as new data is introduced.In traditional training pipelines, the model is continuously trained on a new batch of data, and without a systematic approach, the model may fail to retain knowledge from earlier stages, especially when the data distribution shifts significantly. This results in performance instability when the model encounters tasks that require integrating knowledge from different domains.

The high-certainty pre-training procedure mitigates this issue by ensuring that knowledge is retained across stages. At the end of each training phase, the model’s best-performing checkpoint is selected through the offline decay fitting and weight averaging strategy described earlier. This checkpoint then serves as the initialization for the next stage of training.By systematically selecting and accumulating optimal checkpoints across stages, this process not only prevents catastrophic forgetting but also allows the model to seamlessly integrate knowledge from heterogeneous data sources. This results in a stable and progressive accumulation of capabilities across different domains.In addition to improving optimization stability, the high-certainty procedure also simplifies large-scale training pipelines, making the pre-training process more robust when scaling to heterogeneous datasets and extended training schedules.

\section{Safety-aware Post-training}

The post-training stage aims to align model behavior with human preferences and safety requirements while activating the capabilities learned during pre-training. For JT-Safe-V2, we design a multi-stage alignment pipeline that jointly optimizes general intelligence and safety-by-design principles.

Unlike conventional alignment pipelines that primarily focus on preference learning, the JT-Safe-V2 post-training process emphasizes reliability, contextual reasoning, and safety-aware responses. The alignment strategy consists of two major stages: supervised fine-tuning (SFT) and reinforcement learning (RL). The SFT stage focuses on activating domain capabilities and contextual knowledge through self-distillation and metadata augmentation, while the RL stage further strengthens safety compliance and response quality using structured reward mechanisms.This staged alignment process ensures that the model not only produces helpful responses but also maintains strong safety guarantees when operating in complex real-world environments.

\subsection{Supervised Fine-tuning with Self-Distillation}
Supervised fine-tuning is a critical stage for aligning model behavior and activating domain capabilities. The effectiveness of SFT largely depends on the quality of instruction data and its compatibility with knowledge learned during pre-training. To improve the quality of training signals, we develop a model-driven self-distillation framework that refines instruction datasets through iterative evaluation and filtering.

The process begins with a seed cold-start dataset consisting of approximately on the order of $10^5$ high-quality instruction samples. These samples cover multiple capability domains, including mathematical reasoning, logical reasoning, coding, complex question answering, and instruction following. The goal of this stage is to establish basic conversational abilities and instruction-following patterns.

After the initial model is obtained, reinforcement learning is used to evolve a stronger expert model. This model demonstrates improved stability in understanding complex instructions and generating high-quality responses. The expert model then acts as a teacher for refining the SFT dataset.For each instruction sample, the expert model generates multiple candidate responses. The correctness frequency among these responses is used to estimate the difficulty and reliability of the sample.

Samples for which the expert model fails to produce correct responses are treated as noise or potential annotation errors and are removed from the final dataset. Samples with intermediate difficulty are identified as capability boundary samples and are upsampled to improve the distribution of training difficulty levels.

Through this difficulty-aware refinement process, the final SFT dataset achieves a better balance between reliability and capability expansion.

\subsection{Prefix-Guided Meta-Information Activation of DWC Knowledge}
%

\begin{figure}[ht]
	\centering
	\includegraphics[width=17cm]{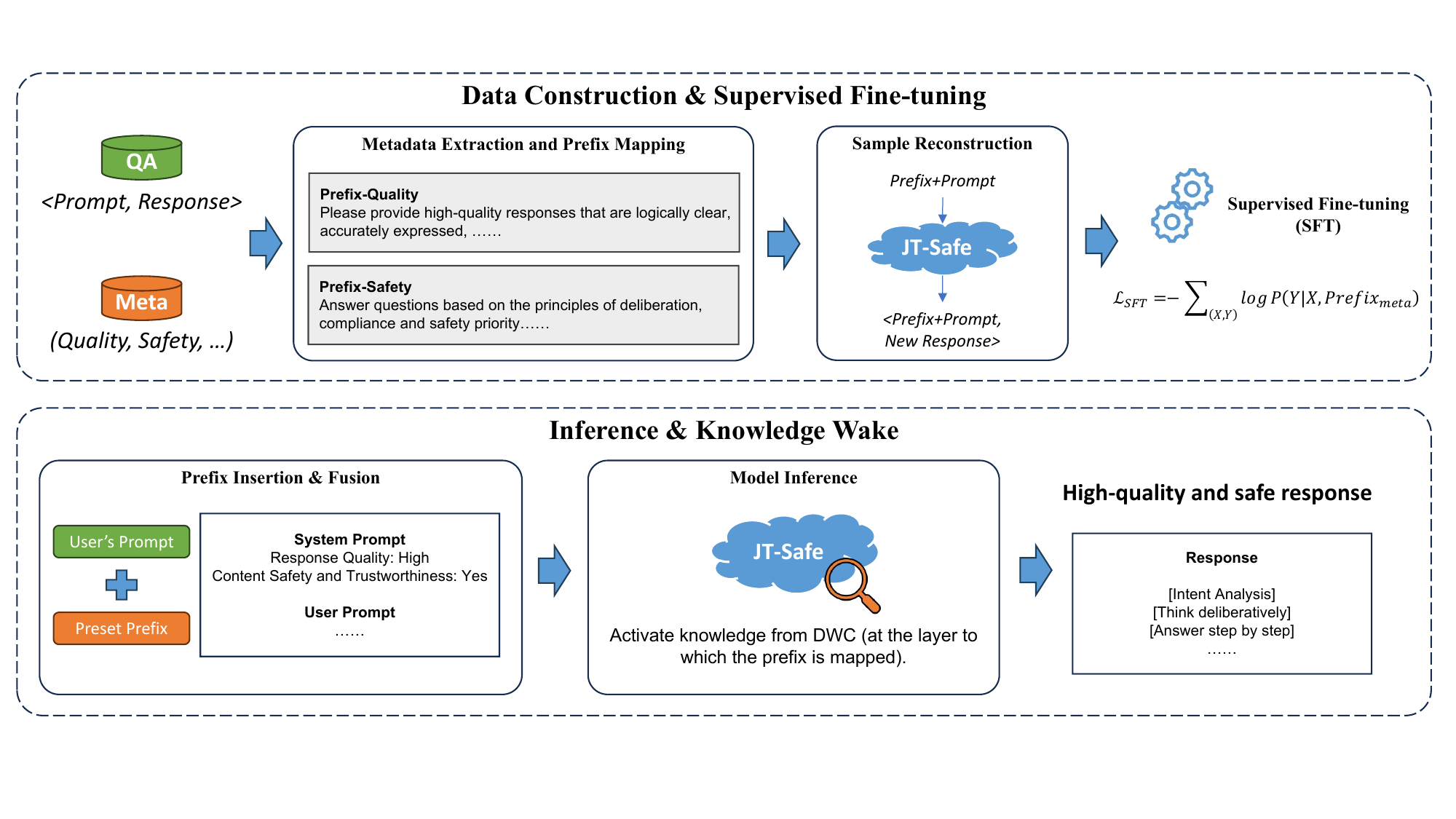}
	\caption{Framework of Prefix-Guided Meta-Information Activation of DWC Knowledge.This figure illustrates our two-stage prefix activation framework. During training, we map quality/safety meta-constraints to prefixes and fine-tune the model; during inference, prefix injection guides the model to activate latent DWC knowledge for high-quality, safe responses.}
	\label{fig:prefix_post_train_fig}
\end{figure}

While the DWC framework integrates extensive contextual world knowledge during pre-training, the model often fails to autonomously elicit latent semantic backgrounds when processing concise user instructions during inference, resulting in performance degradation. To address this challenge, we introduce a prefix activation strategy. This approach employs post-training fine-tuning to enable the model to associate and retrieve latent pre-trained knowledge even from sparse user queries.

The core methodology involves embedding meta-information prefixes within the system prompt to represent specific stylistic and qualitative preferences, which directs the model to activate pertinent internal knowledge. This framework comprises two primary stages: Data Construction and Supervised Fine-tuning, followed by Inference and Knowledge Awakening. Using quality and safety as primary dimensions, prefix activation internalizes the associative mapping between specific triggers and high-quality, trustworthy outputs.

During the Data Construction and Supervised Fine-tuning stage, high-quality instructions are curated from a diverse global dataset. Dual-dimensional constraints encompassing quality, such as logical rigor and professional phrasing, and safety, including harmlessness and compliance, are operationalized into textual prefixes. For example, the prefix may instruct the model to provide responses based on principles of deliberation and safety priority. By concatenating this prefix with original prompts and filtering outputs from a distilled teacher model, a refined dataset of prefix-prompt and response pairs is constructed to meet exceptional quality and safety standards. Subsequently, the model undergoes Supervised Fine-tuning, hereafter SFT, to optimize the conditional probability of the response given the meta-prefix and the prompt. This process compels the model to recognize that specific meta-prefixes require decoding from the distribution of high-quality vocabulary and safe boundary knowledge acquired during pre-training.

During the Inference and Knowledge Awakening stage, providing the model with default meta-guidance shifts the generation process from unconstrained divergence to controlled synthesis. The prefix triggers specific neural pathways established during SFT, prompting the model to proactively mobilize its underlying knowledge network to address queries safely and professionally.

\subsection{Reinforcement Learning Alignment}
After supervised fine-tuning, reinforcement learning is applied to further align model behavior with safety requirements and task objectives. The RL training stage in JT-Safe-V2 is built upon the GRPO algorithm with several strategic modifications designed to improve training stability and performance.To prevent performance stagnation during training, the reference model is periodically updated. This strategy ensures that the optimization process continuously tracks the evolving capability of the model.To mitigate instability caused by extreme off-policy divergence, we adopt conservative clipping coefficients and apply masking to samples that exhibit excessive distributional deviations. These mechanisms improve the robustness of policy updates.

Additionally, we introduce a dynamic temperature modulation mechanism to balance exploration and exploitation during training. By periodically adjusting the sampling temperature, the output entropy is maintained within a desirable range throughout the training process.The training curriculum is also adaptively updated as the model capability improves, ensuring that the model continues to learn from progressively more challenging scenarios.

To guide reinforcement learning optimization, we design a comprehensive reward system that evaluates model outputs across four dimensions: factual accuracy, safety compliance, format consistency, and helpfulness.For factuality evaluation, we introduce a retrieval-augmented verification mechanism. Instead of relying solely on generative judges, the system retrieves supporting evidence from training corpora and trusted websites. The model’s output is compared against the retrieved evidence to estimate factual correctness.Safety evaluation is based on a multi-level safety rubric that defines explicit risk categories and behavioral constraints. A stronger master model is used as a judge to evaluate whether generated responses violate safety guidelines. Violations result in explicit penalties during training.

To address generation artifacts such as language mixing and repetition, a format reward mechanism is introduced. The system checks the consistency between prompt language and output language and detects repetitive patterns near the end of sequences.Finally, helpfulness rewards are implemented through task-specific validators. Mathematical answers are verified using symbolic comparison, coding tasks are evaluated through sandbox execution, and general tasks are assessed through semantic consistency checks.Together, these reward signals ensure that the model generates responses that are not only correct but also safe, coherent, and useful.

\section{Safe-MoMA Framework}

\subsection{Motivation for Multi-Model Orchestration}
As large language models continue to scale in both parameter size and capability, deploying a single monolithic model for all tasks becomes increasingly inefficient. Different models exhibit diverse strengths across task domains such as reasoning, coding, knowledge retrieval, and natural language generation. At the same time, real-world tasks often require structured workflows involving multiple steps, intermediate reasoning, and tool interactions.In enterprise scenarios, user requests are highly heterogeneous. Some queries can be efficiently handled by lightweight models, while others require large reasoning models or specialized agents. Using a single large model for all tasks therefore leads to unnecessary computational overhead and limits system scalability.

To address these challenges, we propose Safe-MoMA (Safe Mixture of Models and Agents), a task-driven orchestration framework that dynamically coordinates multiple models and agents. The framework enables the system to select the most suitable component for each task while maintaining strong safety guarantees.Safe-MoMA is built upon the safety-enhanced JT-Safe-V2 foundation models described in previous sections. By combining safe foundation models with intelligent orchestration, the framework enables efficient, traceable, and safety-aware inference across diverse task scenarios.

\begin{figure}[ht]
	\centering
	\includegraphics[width=17cm]{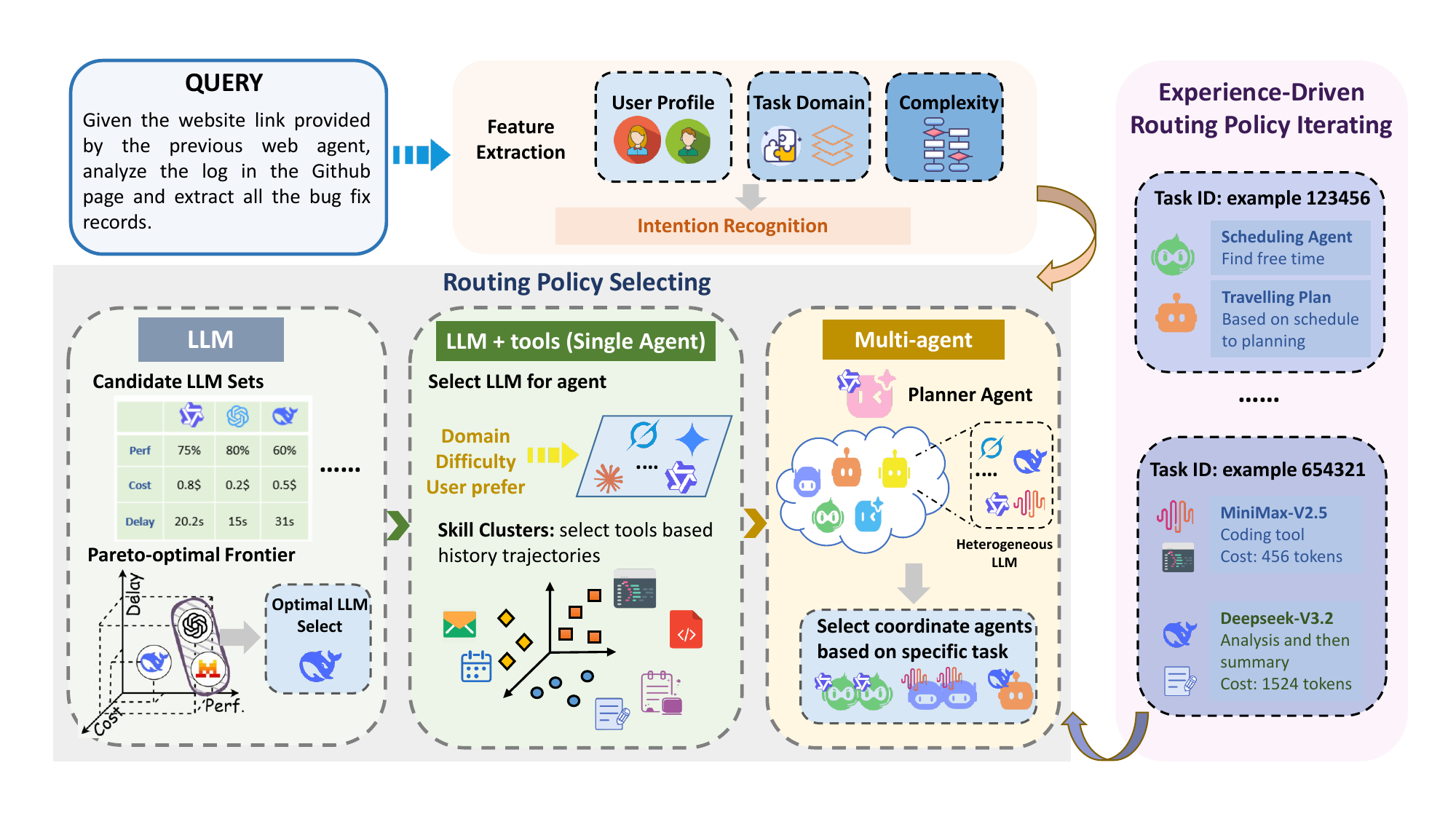}
	\caption{The Adaptive Orchestrator Framework.}
	\label{fig:moma_fig}
\end{figure}

Our method consists of two main components:
\begin{itemize}
	\item \textbf{Capability Boundary Discovery for Models and Agents}, which characterizes the strengths and limitations of different models and agents across task domains and difficulty levels.
	
	\item \textbf{Unified Orchestration Policy Learning}, which formulates the orchestration process as a sequential decision-making problem and learns an optimal orchestration policy via reinforcement learning.
\end{itemize}

\subsection{Capability Boundary Discovery }

In heterogeneous environments with multiple models and tools, different models exhibit varying strengths in terms of reasoning capability, efficiency, and cost. Understanding these capability boundaries is crucial for effective orchestration.
We construct a task collection $D$ from two primary sources: Benchmark datasets and Agent trajectory datasets.
We collect tasks from existing public benchmarks covering diverse domains such as complex reasoning, code generation, tool-use tasks etc. These datasets provide a broad distribution of task difficulty and domains.
We further collect execution trajectories from existing agent systems, including tool invocation sequences, intermediate reasoning steps and final task outcomes. These trajectories capture realistic decision-making processes in agent environments.

We evaluate multiple models on dataset $D$ and measure performance variance across models.
For each task $x_i$, we compute:
\begin{equation}
	\Delta(x_i) =\max_m P(m,x_i) -\min_m P(m,x_i)
\end{equation}  
where $P(m,x_i)$ denotes the performance of model (m) on task $x_i$.
Tasks with large performance variance typically lie near capability boundaries. These tasks are selected as seed tasks.


Using the seed tasks, we employ a strong language model to generate additional task instances with controlled variations $x' \sim G(x_{seed})$.
The generation process varies multiple dimensions, including task difficulty, domain,  reasoning depth and tool dependency. This results in a boundary dataset $D_{boundary}$.


For each model or agent $a$, we estimate a capability function: $C_a(d,l)$, where $d$ denotes task domain and $l$ denotes task difficulty. 
This function estimates the expected performance of an agent under different task conditions.
These capability priors provide important signals for orchestration policy learning.

\begin{table}[t]
	\centering
	\caption{Preferred model or agent for different task types.}
	\label{tab:model_selection}
	\begin{tabular}{ll}
		\toprule
		\textbf{Preferred Model / Agent} & \textbf{Task Type} \\
		\midrule
		DeepSeek-V3.2 & High-complexity reasoning and logical analysis \\
		Qwen3.5-397B-FP8 & Creative generation and multi-turn collaborative content creation \\
		Step3p5-flash & Lightweight text generation and basic question answering \\
		Kimi-K2-5-Thinking & Front-end code generation and interface development \\
		GLM-5 & Backend system development and long-form code generation \\
		MiniMax-V2.1 & Office document understanding and processing (Office / PDF) \\
		Migu\_music & Music playback and generation \\
		migu\_vedio & Video playback and narration \\
		\bottomrule
	\end{tabular}
\end{table}

\subsection{Orchestrator Design Based on Reinforcement Learning}

After constructing capability priors for different models and agents, the next challenge is to dynamically determine how a task should be solved in a heterogeneous environment. In particular, the system must decide whether a task can be directly solved by a single model, requires a single agent with tool usage, or needs a multi-agent collaboration structure. Furthermore, when multi-agent collaboration is necessary, the orchestrator must determine subagents design, including role-playing, models and invoking tools.

To address this problem, we design a task-driven orchestrator and formulate the orchestration process as a sequential decision-making problem. The orchestrator learns a policy that dynamically constructs agent structures and allocates model resources while balancing task performance, computational cost, and inference latency.

\subsubsection{Problem Formulation}

We formulate the orchestration process as a Markov Decision Process (MDP):
\begin{equation}
	\mathcal{M} = (S, A, P, R), 
\end{equation}
where $S$ denotes the state space, $A$ denotes the action space, $P$ represents the transition dynamics, and $R$ is the reward function.

At time step $t$, the system state is defined as $s_t = (x, h_t, r_t, c)$, where:
\begin{itemize}
	\item $x$ is the task description.
	\item $h_t$ represents the execution history up to step $t$, including reasoning traces and tool invocation records.
	\item $r_t$ denotes the current resource usage state, such as accumulated cost and latency.
	\item $c$ represents capability priors of available models and agents derived from the capability boundary discovery stage.
\end{itemize}

These components provide the orchestrator with both task-level information and system-level constraints. The orchestrator selects actions that determine the system execution structure. Actions space mainly include:
\begin{itemize}
	\item selecting the solving paradigm (single model, single agent, or multi-agent).
	\item choosing the model assigned to an agent.
	\item invoking external tools.
\end{itemize}

Formally, an action at time $t$ is represented as $a_t \in A$, which corresponds to a specific orchestration decision.
After executing an action $a_t$, the system transitions to a new state $s_{t+1} \sim P(s_{t+1}|s_t, a_t)$. The transition reflects updates in reasoning history, agent states, and accumulated resource consumption.
The orchestrator receives a reward signal based on task outcome and system efficiency.

\subsubsection{Hierarchical Orchestration Policy}

Directly optimizing a flat action space for orchestration can be challenging due to the combinatorial explosion of possible agent configurations and model assignments. To address this issue, we introduce a hierarchical orchestration policy. The orchestration policy is decomposed into two levels:

\textbf{High-level routing policy}. The first level determines the overall task-solving paradigm:
\begin{equation}
	a^{route} \in
	\{SingleModel, SingleAgent, MultiAgent\}.
\end{equation}

This decision determines whether the task can be solved through direct model inference, requires an agent with tool usage, or needs a collaborative multi-agent workflow. 

\textbf{Low-level orchestration policy}. If the routing policy selects the multi-agent paradigm, the system enters a second decision stage that constructs the agent system dynamically.
The low-level policy determines:  which agent roles to instantiate,  which models to assign to each agent, and which tools should be invoked.
The action at this level is defined as $ a_t = (agent_i, model_j, tool_k)$, where $agent_i$ represents the selected agent role, $model_j$ denotes the assigned model, and $tool_k$ represents the invoked tool.
Through iterative decisions, the orchestrator constructs a dynamic agent workflow: $A = \{a_1, a_2, ..., a_n\}$, where $A$ denotes the instantiated agent set for the current task.
This hierarchical design significantly reduces the decision complexity and improves policy learning stability.

In the hierarchical orchestration policy, the system first determines the execution paradigm for solving a task. Since these paradigms differ significantly in execution structure and resource consumption, the reward calculation must account for these differences while maintaining a unified training objective.

\subsubsection{Reward Design for Hierarchical Orchestration}

In the hierarchical orchestration policy, the system first determines the execution paradigm for solving a task. As illustrated above, we consider three execution modes: \emph{single-model inference}, \emph{single-agent execution}, and \emph{multi-agent collaboration}. Since these paradigms differ significantly in execution structure and resource consumption, the reward calculation must account for these differences while maintaining a unified training objective.

To unify the reward formulation across different paradigms, we introduce an execution indicator variable
\[
z = (z_{SM}, z_{SA}, z_{MA}),
\]
where $z_{SM}$, $z_{SA}$, and $z_{MA}$ indicate whether the task is solved using single-model inference, single-agent execution, or multi-agent collaboration, respectively. The indicator satisfies $z_{SM} + z_{SA} + z_{MA} = 1$, meaning that only one execution paradigm is activated for each task.

A key challenge in multi-agent orchestration is balancing task performance with system efficiency. Larger models typically yield better performance but incur higher computational cost and longer inference latency.
To capture this trade-off, we design a multi-objective reward function.
For each execution paradigm $m \in \{SM, SA, MA\}$, the reward consists of three components:
\[
R^{(m)} = R_{task}^{(m)} - \lambda_c C^{(m)} - \lambda_l L^{(m)}, 
\]
where $R_{task}^{(m)}$ denotes the task completion reward, $C^{(m)}$ represents the computational cost, $L^{(m)}$ denotes the total inference latency, $\lambda_c$ and $\lambda_l$ are weighting coefficients controlling the cost and latency penalties.
Consistent with OpenRouter, the cost of LLM API invocation is conventionally priced per million tokens (differentially for prompt and completion), whereas latency performance is characterized by metrics including Time to First Token (TTFT) and token generation throughput.

\paragraph{Task Reward under Different Execution Paradigms}

Although the overall reward structure remains consistent, the task reward $R_{task}^{(m)}$ differs depending on the execution paradigm.

\begin{itemize}
	\item {Single-model inference and Single-agent execution.}
	In these two mode, the system directly invokes a model or an agent to generate the final output. The task reward therefore depends solely on the final task outcome:
	\[
	R_{task}^{(SM)}  = R_{final}
	\]
	\[
	R_{task}^{(SA)}  = R_{final},
	\]
	where $R_{final}$ denotes the task success reward.
	
	\item {Multi-agent collaboration.}
	In the multi-agent paradigm, tasks are typically decomposed into subtasks solved collaboratively by multiple agents. Therefore, the task reward includes both global task success and subtask-level rewards:
	\[R_{task}^{(MA)} =R_{final}+\beta \sum_k R_{subtask}^{(k)},
	\]
	where $R_{final}$ measures the success of the overall task, $R_{subtask}^{(k)}$ denotes the reward for completing subtask $k$, which is highly related to current LLM selection and tool invocation.
	
\end{itemize}

\paragraph{Unified Reward Formulation}
Combining the execution indicator with the paradigm-specific rewards, the final unified reward function is defined as
\[R =\sum_{m \in \{SM,SA,MA\}}z_m \left(R_{task}^{(m)}-\lambda_c C^{(m)}-\lambda_l L^{(m)}\right).\]

This unified formulation allows the orchestrator to compare different execution strategies under a consistent optimization objective. By jointly considering task success, computational cost, and inference latency, the policy learns to dynamically select the most efficient execution paradigm and resource allocation strategy for each task.

\section{Evaluation}

\subsection{Safety Performance on Comprehensive Benchmarks}

We evaluate the safety performance of \textbf{JT-Safe-V2-35B} across 20 benchmarks spanning six categories, including core values alignment, toxicity and harmful content, bias and fairness, adversarial robustness, safety knowledge, and trustworthiness. We compare against several state-of-the-art models, including Qwen3-32B, Qwen3-235B-A22B, Qwen3.5-35B-A3B, and DeepSeek-V3.2. The benchmarks include:CNsafe~\citep{CNSafe}, CValues~\citep{Cvalues}, Flames~\citep{Flames}, Sweeval~\citep{SweEval}, Airbench~\citep{Air-bench}, AgentHarm~\citep{Agentharm}, SafetyPrompts~\citep{SafetyPrompts}, JADE~\citep{JADE}, SimpleSafetyTest~\citep{Simplesafetytests}, Forbidden~\citep{ForbiddenQuestions}, DoNotAnswer~\citep{DoNotAnswer}, TechHazardQA~\citep{TechHazardQA}, SaladBench~\citep{Salad-bench},
BBQ~\citep{BBQ}, JailbreakBench~\citep{Jailbreakbench}, StrongReject~\citep{StrongReject}, Jbdistill~\citep{JbdiStill}, CSSbench~\citep{CssBench}, ChineseSafetyQA~\citep{ChineseSafetyQA}, TruthfulQA~\citep{TruthfulQA}.

\paragraph{Overall Performance.}
Across all benchmarks, JT-Safe-V2-35B demonstrates strong and stable performance. It achieves competitive or leading results on most tasks and maintains consistent behavior across different safety dimensions. Notably, JT-Safe-V2-35B achieves a near-perfect score of 100.00 on the CValues benchmark, and remains highly competitive on others like CNsafe (99.66). Compared with baseline models, such as Qwen3-32B, JT-Safe-V2-35B exhibits better overall balance and reliability, indicating the effectiveness of its safety-by-design training paradigm.

\paragraph{Core Values Alignment.}
On core value alignment benchmarks (CNsafe, CValues, and Flames), JT-Safe-V2-35B achieves near-saturated performance. It obtains the highest score on CValues (100.00) and remains highly competitive on the other benchmarks, scoring 99.66 on CNsafe and 99.44 on Flames. These results suggest that the model effectively captures normative and value-aligned behaviors, likely benefiting from the structured signals introduced in the DWC framework.

\paragraph{Toxicity and Harmful Content.}
On the 10 benchmarks related to toxicity and harmful content, JT-Safe-V2-35B shows consistently strong performance. It achieves top results on Sweeval (95.85) and SafetyPrompts (98.92), and maintains near-perfect scores on SimpleSafetyTest (100.00) and Forbidden (99.86). Compared with smaller models such as Qwen3-32B, JT-Safe-V2-35B shows substantial improvements, especially on more challenging datasets like Sweeval. In addition, it avoids large performance fluctuations observed in some models, indicating more stable safety behavior across different risk scenarios.

\paragraph{Bias and Fairness.}
On the BBQ benchmark, all models achieve similar performance, generally around 94--95. JT-Safe-V2-35B achieves the highest score (94.91), although the margin is relatively small. This suggests that bias mitigation remains a challenging and relatively saturated problem, and further improvements may require more targeted data and training strategies.

\paragraph{Adversarial Robustness.}
Adversarial robustness is a key dimension in safety evaluation. JT-Safe-V2-35B performs strongly on these benchmarks. It achieves the best result on StrongReject (99.79) and remains competitive on JailbreakBench (99.30) and CSSbench (96.76). Compared with Qwen3-32B, JT-Safe-V2-35B shows significant improvements on Jbdistill. Although some models achieve higher scores on specific benchmarks, their performance is less consistent. JT-Safe-V2-35B provides a more balanced robustness profile across different attack settings.

\paragraph{Safety Knowledge.}
On ChineseSafetyQA, JT-Safe-V2-35B achieves a clear advantage with a score of 97.76. It outperforms all baseline models by a large margin, especially compared with smaller models like Qwen3-32B, which scores 81.33. This indicates strong internalization of safety-related knowledge, likely due to the factual and cognitive annotations introduced during pre-training.

\paragraph{Trustworthiness.}
On TruthfulQA, JT-Safe-V2-35B achieves the highest score (81.57). Although the absolute performance remains lower compared to other categories, reflecting the difficulty of the task, the improvement suggests better control of hallucination and more reliable responses. It also performs strongly on RagTruth (92.17), further supporting the model's reliability.

Overall, JT-Safe-V2-35B achieves a strong combination of safety, robustness, and consistency across diverse benchmarks. Compared with existing models, it provides more stable cross-domain safety performance, stronger adversarial robustness, and clear advantages in safety knowledge and trustworthiness. These results demonstrate that the proposed safety-by-design training paradigm, together with structured DWC data, can effectively improve model safety without sacrificing overall performance.Unlike conventional training data that primarily relies on raw text, DWC introduces structured annotations across factual, logical, and cognitive layers, providing the model with more explicit supervision signals.Specifically, the factual layer enhances grounding and reduces ambiguity, which contributes to improved performance on safety knowledge and truthfulness benchmarks. The logical layer exposes reasoning structures and causal relationships, enabling the model to better identify unsafe intent and resist adversarial prompts. The cognitive layer encodes value alignment, intent awareness, and audience adaptation, which supports consistent performance on core value alignment and toxicity-related tasks.By jointly integrating these three layers, DWC allows the model to learn not only what to say, but also how and why to respond in a safe manner. This structured supervision leads to more stable and generalizable safety behavior across diverse scenarios, demonstrating that data-centric design is a key factor in achieving safety-by-design foundation models.

\begin{table*}[t]
	\centering
	\small
	\setlength{\tabcolsep}{4pt}
	\caption{Safety performance across 20 benchmarks grouped into six categories. Higher is better.}
	\label{tab:safety_results}
	\begin{tabular}{llccccc}
		\toprule
		\textbf{Category} & \textbf{Benchmark} & \textbf{Qwen3-32B} & \textbf{Qwen3-235B} & \textbf{Qwen3.5-35B} & \textbf{DeepSeek-V3.2} & \textbf{JT-Safe-V2-35B} \\
		\midrule
		
		\multirow{3}{*}{Core Values}
		& CNsafe & 95.43 & 99.61 & 99.55 & 97.62 & \textbf{99.66} \\
		& CValues & 99.75 & 99.94 & 99.96 & 99.86 & \textbf{100.00} \\
		& Flames & 96.78 & \textbf{99.50} & \textbf{99.50} & 98.86 & 99.44 \\
		
		\midrule
		
		\multirow{10}{*}{\makecell[l]{Toxicity \&\\Harmful Content}}
		& Sweeval & 64.26 & 92.51 & 92.63 & 80.53 & \textbf{95.85} \\
		& Airbench & 87.81 & 97.00 & \textbf{98.59} & 94.95 & 97.37 \\
		& AgentHarm & 92.77 & 94.62 & 84.23 & 90.00 & \textbf{94.69} \\
		& SafetyPrompts & 92.53 & 95.75 & 98.38 & 95.58 & \textbf{98.92} \\
		& JADE & 98.37 & 99.73 & \textbf{99.97} & 99.59 & 99.76 \\
		& SimpleSafetyTest & 99.88 & 99.52 & \textbf{100.00} & 99.28 & \textbf{100.00} \\
		& Forbidden & 97.04 & 99.49 & \textbf{99.86} & 99.58 & \textbf{99.86} \\
		& DoNotAnswer & 99.38 & 99.96 & \textbf{99.98} & 99.84 & 99.91 \\
		& TechHazardQA & 93.29 & 98.65 & 98.14 & 97.70 & \textbf{98.88} \\
		& SaladBench & 86.73 & 91.15 & 88.49 & 89.66 & \textbf{94.10} \\
		
		\midrule
		
		\multirow{1}{*}{Bias \& Fairness}
		& BBQ & 94.40 & 94.76 & 94.76 & 94.30 & \textbf{94.91} \\
		
		\midrule
		
		\multirow{4}{*}{Adversarial Robustness}
		& JailbreakBench & \textbf{99.40} & 97.40 & 96.80 & 90.60 & 99.30 \\
		& StrongReject & 90.64 & 97.58 & 99.01 & 88.13 & \textbf{99.79} \\
		& Jbdistill & 78.42 & 91.15 & \textbf{97.06} & 84.89 & 96.24 \\
		& CSSbench & 81.72 & 93.85 & \textbf{98.22} & 89.18 & 96.76 \\
		
		\midrule
		
		\multirow{1}{*}{Safety Knowledge}
		& ChineseSafetyQA & 81.33 & 87.06 & 85.77 & 88.73 & \textbf{97.76} \\
		
		\midrule
		
		\multirow{2}{*}{Trustworthiness}
		& TruthfulQA & 73.41 & \textbf{81.88} & 79.90 & 79.68 & 81.57 \\
		& RagTruth & 89.04 & 92.09 & 69.61 & 93.92 & \textbf{92.17} \\
		\bottomrule
	\end{tabular}
\end{table*}

\subsection{General Capabilities Evaluation}

To further evaluate whether safety improvements come at the cost of general capabilities, we compare JT-Safe-V2-35B with strong baselines, including Qwen3-235B and other state-of-the-art models of comparable scale. This setting provides a stringent evaluation, as these baselines represent the upper bound of current large-scale models.

Table~\ref{tab:general_capabilities_full} presents the results across seven capability dimensions, including coding(HumanEval~\citep{chen2021humaneval}, HumanEval+ ~\citep{liu2023evalplus}, MBPP+~\citep{liu2023evalplus}, CruxEval~\citep{gu2024cruxeval}, MultiPL-E~\citep{cassano2023multipl}), mathematics(GSM8K~\citep{cobbe2021gsm8k}, MATH500~\citep{lightman2023lets}, MGSM-zh~\citep{shi2022language}, MathBench~\citep{liu2024mathbench}), long-context understanding(MRCR~\citep{vodrahalli2024michelangelolongcontextevaluations}, RULER~\citep{hsieh2024rulerwhatsrealcontext}, Leval~\citep{an2023levalinstitutingstandardizedevaluation}), reasoning(PIQA~\citep{piqa}, HellaSwag~\citep{HellaSwag}, StrategyQA~\citep{strategyQA}, AutoLogi~\citep{zhu2025autologi}), knowledge(CryptoBench~\citep{CryptoBench}, FinEval~\citep{FinEval}, Xiezhi~\citep{Xiezhi}, GaokaoBench~\citep{GaokaoBench}), agent capability(BFCL~\citep{patil2025bfcl}, $\tau^2$-Bench~\citep{barres2025tau2benchevaluatingconversationalagents}, SEAL~\citep{pham2026sealqaraisingbarreasoning}, GAIA-text~\citep{gaia2023}), and low-resource language (SEA-HELM~\citep{susanto2025seahelmsoutheastasianholistic}) performance.

\begin{table*}[t]
	\centering
	\small
	\setlength{\tabcolsep}{5pt}
	\caption{General capabilities results}
	\label{tab:general_capabilities_full}
	\begin{tabular}{llcc}
		\toprule
		\textbf{Category} & \textbf{Benchmark} & \textbf{SOTA with Equivalent Parameters} & \textbf{JT-Safe-V2-35B} \\
		\midrule
		
		\multirow{5}{*}{Coding}
		& HumanEval & 94.50 & 96.30 \\
		& HumanEval+ & 89.00 & 90.24 \\
		& MBPP+ & 80.02 & 78.87 \\
		& CruxEval & 75.70 & 88.50 \\
		& MultiPL-E & 87.90 & 82.47 \\
		
		\midrule
		
		\multirow{3}{*}{Mathematics}
		& GSM8K & 93.93 & 94.62 \\
		& MATH500 & 95.80 & 97.20 \\
		& MGSM-zh & 89.20 & 89.60 \\
		& MathBench & 74.26 & 94.87 \\
		\midrule
		
		\multirow{3}{*}{Long Context}
		
		& MRCR & 15.60 & 21.60 \\
		& RULER & 95.00 & 93.40 \\
		& Leval & 42.80 & 42.60 \\
		\midrule
		
		\multirow{4}{*}{Reasoning}
		& PIQA & 70.46 & 84.22 \\
		& HellaSwag & 65.27 & 87.27 \\
		& StrategyQA & 62.36 & 83.19 \\
		& AutoLogi & 92.31 & 93.02 \\
		
		\midrule
		
		\multirow{4}{*}{Knowledge}
		& CryptoBench & 75.16 & 79.39 \\
		& FinEval & 77.50 & 78.89 \\
		& Xiezhi & 52.35 & 68.02 \\
		& GaokaoBench & 27.76 & 92.28 \\
		
		\midrule
		
		\multirow{4}{*}{Agent}
		& BFCL & 72.90 & 73.69 \\
		& $\tau^2$-Bench & 48.10 & 70.00 \\
		& SEAL & 57.40 & 32.11 \\
		& GAIA-text & 84.50 & 63.11 \\
		
		\midrule
		
		\multirow{1}{*}{Low-resource}
		& SEA-HELM & 60.65 & 65.30 \\
		
		\bottomrule
	\end{tabular}
\end{table*}

\paragraph{Coding ability.}
JT-Safe-V2-35B demonstrates strong coding performance, outperforming the baseline on HumanEval (96.30 vs. 94.50) and CruxEval (88.50 vs. 75.70), while remaining competitive on other benchmarks. Although slightly lower on MultiPL-E, the overall results indicate that safety-oriented training does not degrade code generation ability.

\paragraph{Mathematical ability.}
On mathematical benchmarks, JT-Safe-V2-35B achieves consistent improvements, surpassing the baseline on GSM8K (94.62 vs. 93.93) and MATH500 (97.20 vs. 95.80), while maintaining comparable performance on MGSM-zh. This demonstrates strong reasoning and numerical capability.

\paragraph{Long-context understanding.}
JT-Safe-V2-35B shows significant advantages in long-context tasks, especially on MathBench (94.87 vs. 74.26) and MRCR (21.60 vs. 15.60). Although slightly lower on RULER, the results indicate improved capability in handling complex long-form inputs.

\paragraph{Reasoning ability.}
Substantial gains are observed in reasoning benchmarks. JT-Safe-V2-35B significantly outperforms the baseline on PIQA (84.22 vs. 70.46), HellaSwag (87.27 vs. 65.27), and StrategyQA (83.19 vs. 62.36), demonstrating strong improvements in commonsense and multi-step reasoning.

\paragraph{Knowledge capability.}
On knowledge-intensive tasks, JT-Safe-V2-35B achieves strong performance, particularly on GaokaoBench (92.28 vs. 27.76) and Xiezhi (68.02 vs. 52.35). These results indicate that the model maintains and even improves knowledge-related capabilities despite safety-oriented training.

\paragraph{Agentic capability.}
Results on agent-related benchmarks show mixed performance. JT-Safe-V2-35B improves on $\tau^2$-Bench (70.00 vs. 48.10), indicating stronger planning and reasoning abilities, but lags behind on SEAL and GAIA-text. This suggests that agent capability remains an area for further improvement.

\paragraph{Low-resource language ability.}
On SEA-HELM, JT-Safe-V2-35B achieves a higher score (65.30 vs. 60.65), indicating improved performance in low-resource language scenarios.

\paragraph{Overall analysis.}
Overall, JT-Safe-V2-35B demonstrates strong and well-balanced performance across diverse capability dimensions. Notably, it achieves significant gains in reasoning, knowledge, and long-context tasks, while maintaining competitive performance in coding and mathematics.

These results show that improving safety does not come at the cost of general capabilities. Instead, the combination of self-distillation and DWC-based structured data enhances reasoning, stabilizes knowledge application, and improves long-context understanding. The structured supervision introduced by DWC enables the model to better organize information and produce more coherent outputs, which contributes to improvements across multiple capability dimensions.

\subsection{Ablation Study of DWC across Pre-training and Fine-tuning}

\subsubsection{Impact of DWC Meta-information during Continued Pre-training}

\begin{figure}[ht]
	\centering
	\includegraphics[width=17cm]{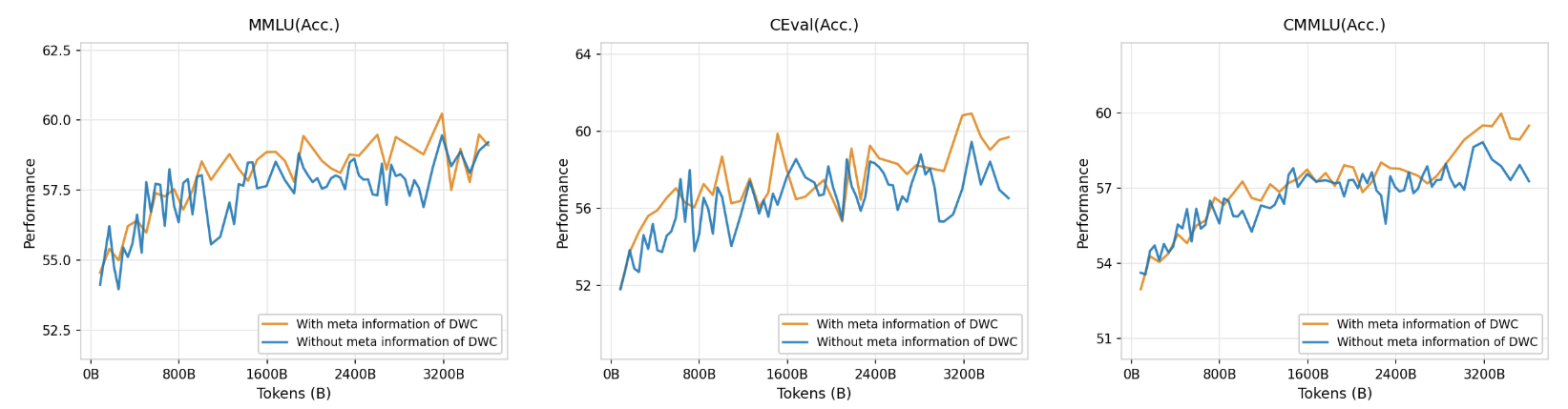}
	\caption{Performance trajectories during continued pre-training with DWC meta-information. The plots compare model performance as a function of training tokens (in billions) across three representative knowledge benchmarks: MMLU, C-Eval, and CMMLU.}
	\label{fig:pretrain_ablation}
\end{figure}

We further conduct an ablation study in the continued pre-training stage to evaluate the effect of introducing DWC meta-information. The experiments are not conducted from scratch, but instead start from a well-trained checkpoint, and continue training with or without meta-information.

Figure \ref{fig:pretrain_ablation}  illustrates the performance trajectories on three representative knowledge-intensive benchmarks, including MMLU, C-Eval, and CMMLU, as a function of training tokens (in billions). Across all three benchmarks, models trained with DWC meta-information consistently outperform those without meta-information throughout the training process.

Notably, the performance gap emerges early and remains stable as training progresses. This indicates that DWC meta-information improves not only the final performance but also the efficiency of knowledge acquisition during continued training. In other words, for the same amount of training data, models with meta-information achieve higher performance, demonstrating improved data efficiency.

An important observation is that the introduction of DWC meta-information exhibits strong plug-and-play compatibility. Since the training starts from an already well-trained checkpoint, the improvement cannot be attributed to re-initialization or large-scale retraining. Instead, it shows that meta-information can be injected at arbitrary stages of the model lifecycle without requiring architectural changes or retraining from scratch.

These results highlight that DWC meta-information serves as a lightweight yet effective enhancement mechanism. By providing structured signals during training, it accelerates knowledge consolidation and improves factual reliability, ultimately leading to more efficient and trustworthy model behavior.

\subsubsection{Effect of DWC Meta-information in Fine-tuning}

\begin{table*}[t]
	\centering
	\small
	\setlength{\tabcolsep}{4pt}
	\caption{Impact of DWC meta-information in the fine-tuning stage. We compare performance with and without meta-information across multiple benchmarks. $\Delta$ denotes the performance gain after introducing meta-information.}
	\label{tab:dwc_meta_ablation}
	\begin{tabular}{lccc ccc ccc ccc}
		\toprule
		
		\textbf{Model} 
		& \multicolumn{3}{c}{\textbf{Chinese-SimpleQA}} 
		& \multicolumn{3}{c}{\textbf{TruthfulQA}} 
		& \multicolumn{3}{c}{\textbf{C-Eval}} 
		& \multicolumn{3}{c}{\textbf{CMMLU}} \\
		
		\cmidrule(lr){2-4} \cmidrule(lr){5-7} \cmidrule(lr){8-10} \cmidrule(lr){11-13}
		
		& w/ meta & w/o meta & $\Delta$
		& w/ meta & w/o meta & $\Delta$
		& w/ meta & w/o meta & $\Delta$
		& w/ meta & w/o meta & $\Delta$ \\
		
		\midrule
		
		JT-Safe-V2-35B
		& 50.46 & 47.93 & +2.53
		& 86.46 & 81.52 & +4.94
		& 87.42 & 87.08 & +0.34
		& 83.28 & 82.93 & +0.35 \\
		
		Qwen3-235B
		& 82.60 & 82.00 & +0.60
		& 86.96 & 87.22 & -0.26
		& 87.45 & 88.99 & -1.54
		& 86.48 & 88.67 & -2.19 \\
		
		Qwen3-32B
		& 45.50 & 44.76 & +0.74
		& 83.67 & 82.41 & +1.26
		& 87.72 & 87.22 & +0.50
		& 83.68 & 83.65 & +0.03 \\
		
		Qwen3-30B
		& 68.83 & 68.63 & +0.20
		& 84.68 & 83.29 & +1.39
		& 85.83 & 84.44 & +1.39
		& 83.94 & 84.03 & -0.09 \\
		
		\bottomrule
	\end{tabular}
\end{table*}

To further investigate the effectiveness of DWC, we conduct controlled experiments during the fine-tuning stage. The only difference between the two settings is whether meta-information is included. Since meta-information is the core component of DWC—comprising factual, logical, and cognitive layers—this setup allows us to directly evaluate its contribution.

Table~\ref{tab:dwc_meta_ablation} presents the results across four knowledge-intensive benchmarks.

For JT-based models, incorporating meta-information leads to consistent improvements across all benchmarks. On Chinese-SimpleQA, performance increases from 47.93 to 50.46 (+2.53), and on TruthfulQA, the gain reaches +4.94, which is the largest improvement among all tasks. More moderate but stable gains are also observed on C-Eval (+0.34) and CMMLU (+0.35). These results indicate that meta-information provides effective supervision signals that enhance knowledge retrieval, factual grounding, and answer reliability.

In contrast, the effect of meta-information on baseline models is inconsistent and sometimes negative. For example, Qwen3-235B shows performance degradation on multiple benchmarks, including C-Eval (-1.54) and CMMLU (-2.19). Similar patterns are observed for other baseline models, where improvements are limited or unstable. This suggests that simply adding meta-information does not guarantee performance gains and may even introduce noise when it is not properly aligned with the model’s training process.

The consistent improvements observed in JT-based models highlight the importance of structured and aligned meta-information in DWC. Specifically, the factual layer enhances grounding and reduces ambiguity, which contributes to the strong gains on Chinese-SimpleQA and TruthfulQA. The logical layer helps organize information and supports more coherent reasoning, while the cognitive layer encodes intent and task-level signals, enabling the model to better interpret and utilize the input.

Overall, these results demonstrate that the three-layer annotation scheme in DWC is both effective and necessary. By integrating factual, logical, and cognitive information in a structured manner, DWC enables the model to leverage meta-information more effectively, leading to stable and consistent improvements across knowledge-intensive benchmarks.

\subsection{Efficiency Comparison of MoMA and Routing Models}
\textbf{Comparison with single LLM:} When evaluating six single models, Qwen3-235B-A22B achieves the highest score (68.6) across three benchmarks. Deepseek-r1 followed closely with 60.2, as shown in Table~\ref{table}.
Compared to a single LLM, MoMA achieves state-of-the-art performance in both AIME2024 and SimpleQA benchmarks under performance-priority scenarios. Compared to the optimal single model (Qwen3-235B-A22B), it achieves comparable performance (with a 2.9\% score improvement) while reducing costs by 31.46\%.

\textbf{Comparison with other routing frameworks: } 
MoMA router with the performance-first preference achieves optimal performance. Its automated routing strategy achieves a relatively high score (surpassing Deepseek-V3) at a significantly lower cost (37.19\% reduction compared to the performance-priority), thereby achieving an optimal trade-off between performance and cost. 
The SFT-based approach, with only an optimizing model as output, fails to achieve a cost-performance trade-off.
Although it performs best under the auto-routing preference across the three routing frameworks, this advantage stems from our relatively constrained data categories, such methods perform well under limited category conditions. However, in practical applications involving numerous categories, its performance degrades significantly. Moreover, its computational cost is higher than the other two auto-routing frameworks, achieving only marginal performance gains.
Contrastive learning-based methods exhibit performance comparable to MoMA, yet MoMA achieves lower computational and training costs among the three preferences.

\begin{table}[htbp]
	\caption{Performance and cost comparison of MoMA with single-model and other routing methods.}
	\centering
	\label{table}
	\resizebox{\textwidth}{!}
	{
		\begin{tabular}{cccccc|c}
			\hline
			\multicolumn{2}{c}{LLMs}                                                                                                                  & AIME2024      & LiveCodeBench & SimpleQA      & Average Score  & Cost \\ \hline
			\multicolumn{2}{c}{Deepseek-R1}                                                                                                           & 79.8          & \textbf{73.1} & 27.8          & 60.2          & 12.327    \\
			\multicolumn{2}{c}{Deepseek-V3}                                                                                                           & 59.4          & 27.2          & 24.9          & 37.2          & 9.498    \\
			\multicolumn{2}{c}{Qwen3-32B}                                                                                                             & 81.4          & 60.7          & 8.0           & 50.0          & 14.65    \\
			\multicolumn{2}{c}{Qwen3-235B-A22B}                                                                                                        & 85.7 & 65.9          & 54.3 & 68.6 & 14.65    \\
			\multicolumn{2}{c}{JT-Math-8B}                                                                                                       & 37.5          & -             & -             & -              & 1.667    \\
			\multicolumn{2}{c}{JT-Code-8B}                                                                                                       & -             & 26.3          & -             & -              & 1.667    \\ \hline
			\multicolumn{1}{c|}{\multirow{3}{*}{\textbf{MoMA Router}}}                                                         & Cost-proirity        & 35.8          & 24.6          & 12.1          & 24.2          & 1.357    \\
			\multicolumn{1}{c|}{}                                                                                              & Auto-routing         & 65.2          & 45.3          & 19.5          & 43.3          & 6.306    \\
			\multicolumn{1}{c|}{}                                                                                              & Performance-priority & \textbf{87.3}          & 66.5          & \textbf{56.3}          & \textbf{70.1 }          & 10.04   \\ \hline
			\multicolumn{1}{c|}{\begin{tabular}[c]{@{}c@{}}SFT-based \\ Classification Router\end{tabular}}                     & Auto-routing         & 76.8          & 70.5          & 40.7          & 62.7          & 8.667    \\ \hline
			\multicolumn{1}{c|}{\multirow{3}{*}{\begin{tabular}[c]{@{}c@{}}Contrastive learning \\ based Router\end{tabular}}} & Cost-proirity        & 31.7          & 27.6          & 14.2          & 24.5           & 1.667    \\
			\multicolumn{1}{c|}{}                                                                                              & Auto-routing         & 65.7          & 40.1          & 17.8          & 41.2           & 6.940    \\
			\multicolumn{1}{c|}{}                                                                                              & Performance-priority & 81.2          & 61.3          & 38.7          & 60.4           & 12.498    \\ \hline
	\end{tabular}}
\end{table}

\section{Conclusion}
This report introduces JT-Safe-V2-35B, a series of foundation models designed to enhance safety and trustworthiness through the joint optimization of general intelligence and safety-by-design. By adopting the Data-with-Context (DWC) paradigm, we integrate structured meta-information across factual, logical, and cognitive layers into pre-training, combined with high-certainty training strategies and safety-enhanced post-training for enterprise-level agentic applications. Extensive experiments show that JT-Safe-V2-35B achieves leading or competitive performance across both safety and general capability benchmarks, demonstrating strong robustness, trustworthiness, and high-level reasoning ability, while maintaining performance comparable to significantly larger models. Ablation studies further confirm that DWC meta-information consistently improves knowledge acquisition efficiency in both fine-tuning and continued pre-training, with strong plug-and-play compatibility. Building on this, we propose Safe-MoMA, a multi-model and multi-agent framework that enables efficient and traceable inference, reducing inference cost by more than 30\% while preserving performance. Overall, this work highlights that model safety should be achieved through a unified design that integrates rich world knowledge and safety considerations across the entire training lifecycle, offering a scalable and practical pathway toward safety-by-design foundation models.

\bibliographystyle{unsrtnat}
\bibliography{references}  

\clearpage

\end{document}